%% file: main.tex
\documentclass[letterpaper,10pt,conference]{ieeeconf}

\IEEEoverridecommandlockouts %
\newcommand{\ELBO}{\mathrm{ELBO}}
\newcommand{\BCE}{\mathrm{BCE}}
\newcommand{\MSE}{\mathrm{MSE}}
\newcommand{\enc}{\mathrm{enc}}
\newcommand{\dec}{\mathrm{dec}}

\newcommand{\T}{\top} %

\usepackage[utf8]{inputenc}

\usepackage{cite}  %
\usepackage{graphics}
\usepackage{epsfig}
\usepackage{times}
\usepackage{amsmath}
\usepackage{amssymb}
\usepackage{subcaption}
\usepackage[bookmarks=false]{hyperref}
\usepackage{siunitx}
\usepackage[ruled,vlined]{algorithm2e}

\usepackage{flushend} %

\input{math_commands}

\begin{document}
\bstctlcite{IEEEexample:BSTcontrol} 

\title{\LARGE\bf
    Next Steps: Learning a Disentangled Gait Representation \\ for Versatile Quadruped Locomotion
}

\author{
    Alexander L. Mitchell, Wolfgang Merkt, Mathieu Geisert, Siddhant Gangapurwala, \\ 
    Martin Engelcke, Oiwi Parker Jones, Ioannis Havoutis, and Ingmar Posner%
    \thanks{All authors are with the Oxford Robotics Institute, University of Oxford}%
    \thanks{Email: \texttt{mitch@robots.ox.ac.uk}}%
    \thanks{This work was supported by a UKRI/EPSRC Programme Grant [EP/V000748/1], the EPSRC grant `Robust Legged Locomotion' [EP/S002383/1], the EPSRC CDT [EP/L015897/1], the UKRI/EPSRC RAIN [EP/R026084/1] and ORCA [EP/R026173/1] Hubs and the EU H2020 Project MEMMO (780684). It was conducted as part of ANYmal Research, a community to advance legged robotics.}
}

\thispagestyle{empty}
\pagestyle{empty}

\maketitle

\input{0_abstract}

\IEEEpeerreviewmaketitle

\input{1_intro}

\input{2_related_work}

\input{3_methods_section}

\input{4_experiments}

\input{5_conclusions}

\input{6_appendix}

\bibliographystyle{IEEEtran}
\bibliography{references}

\end{document}

%% file: math_commands.tex
\usepackage{amsmath,amsfonts,bm}

\def\eqref#1{equation~\ref{#1}}

\def\1{\bm{1}}

\def\rva{{\mathbf{a}}}

\def\rvc{{\mathbf{c}}}

\def\rvq{{\mathbf{q}}}

\def\rvs{{\mathbf{s}}}

\def\rvx{{\mathbf{x}}}

\def\rvz{{\mathbf{z}}}

\def\rmC{{\mathbf{C}}}

\def\rmI{{\mathbf{I}}}

\def\rmQ{{\mathbf{Q}}}

\def\rmS{{\mathbf{S}}}

\def\rmX{{\mathbf{X}}}

\DeclareMathAlphabet{\mathsfit}{\encodingdefault}{\sfdefault}{m}{sl}
\SetMathAlphabet{\mathsfit}{bold}{\encodingdefault}{\sfdefault}{bx}{n}

\newcommand{\KL}{D_{\mathrm{KL}}}

%% file: 0_abstract.tex
\begin{abstract}

Quadruped locomotion is rapidly maturing to a degree where robots now routinely traverse a variety of unstructured terrains. However, while gaits can be varied typically by selecting from a range of pre-computed styles, current planners are unable to vary key gait parameters \emph{continuously} while the robot is in motion. The synthesis, on-the-fly, of gaits with unexpected operational characteristics or even the blending of dynamic manoeuvres lies beyond the capabilities of the current state-of-the-art. In this work we address this limitation by learning a latent space capturing the key stance phases of a particular gait, via a generative model trained on a single trot style.
This encourages disentanglement such that application of a \emph{drive signal} to a single dimension of the latent state induces holistic plans synthesising a continuous variety of trot styles.
In fact properties of this drive signal map directly to gait parameters such as cadence, footstep height and full stance duration.
The use of a generative model facilitates the detection and mitigation of disturbances to provide a versatile and robust planning framework. We evaluate our approach on a real ANYmal quadruped robot and demonstrate that our method achieves a continuous blend of dynamic trot styles whilst being robust and reactive to external perturbations.

\end{abstract}

%% file: 1_intro.tex
\section{Introduction}
\begin{figure}[tb]
    \centering
    \includegraphics[width=1.0\linewidth]{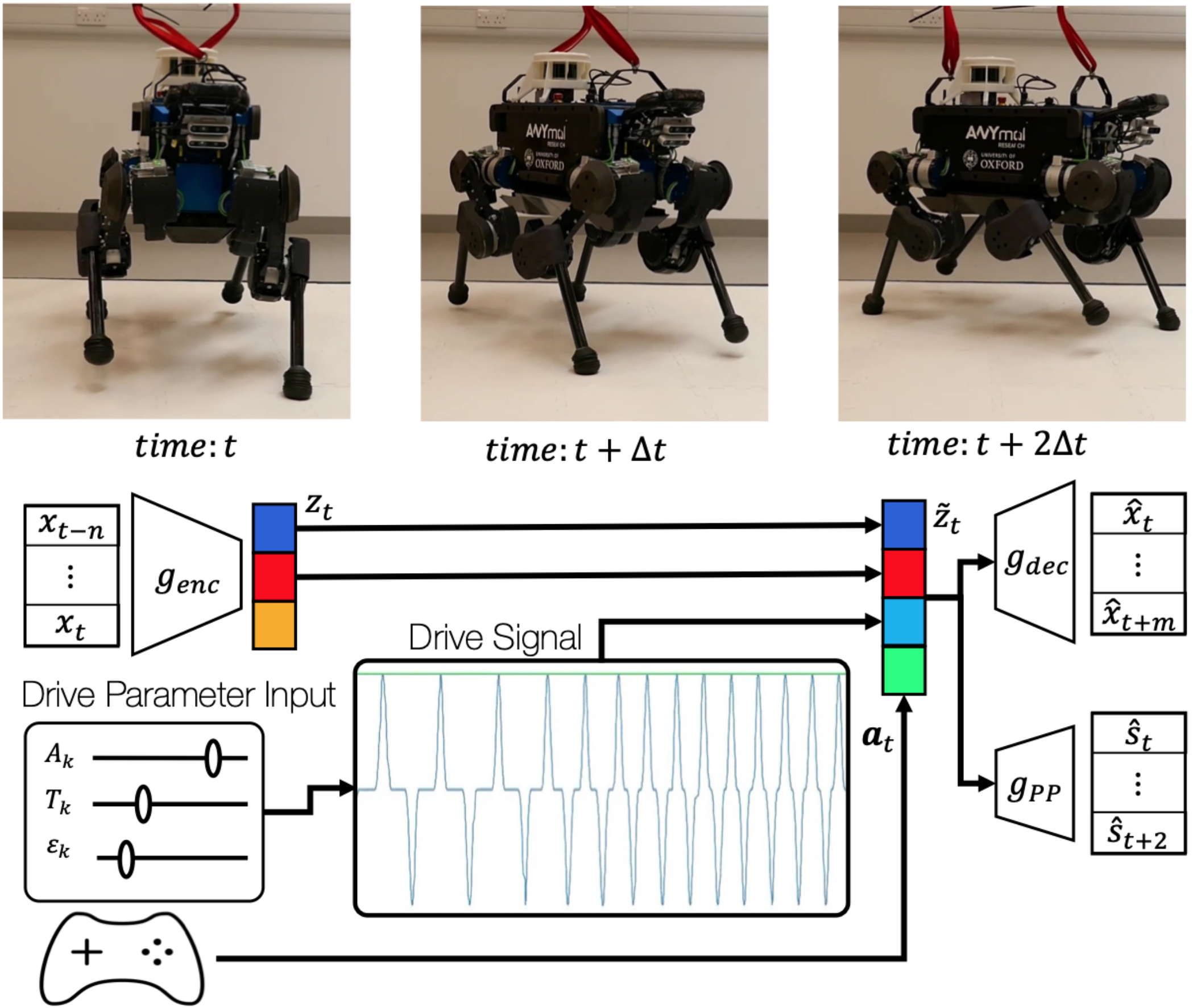}
    \caption{
    Using a variational auto-encoder (VAE), our approach learns a structured latent space capturing key stance phases constituting a particular gait. The space is disentangled to a degree such that application of a \emph{drive signal} to a single dimension of the latent variable induces gait styles which can be seamlessly interpolated between. 
    We encode raw sensor information to infer the robot's gait phase using $g_{\enc}$ before applying our drive signal and then decode the augmented latent variable and the base twist action $\rva_k$ via $g_{\dec}$ and predict the feet in contact using $g_{pp}$. 
    The drive signal's amplitude and phase provide continuous control over the robot's cadence, full-support duration and foot swing height.
    }
    \label{fig:first_figure}
    \vspace{-0.5cm}
\end{figure}

Quadruped locomotion has advanced significantly in recent years, extending its capability towards applications of significant value to industry and the public domain. Driven primarily by advances in optimisation-based~\cite{bellicoso2018dynamic,boxFDDP,melon2021receding,towr} and reinforcement learning-based methods~\cite{hwangbo2019learning,gcpo,gangapurwala2020rloc}, quadrupeds are now able to traverse over a wide variety of terrains, making them a popular choice for tasks such as inspection, monitoring, search and rescue or goods delivery in difficult, unstructured environments. However, despite recent advances, important limitations remain. Due to the complexity of the system, models used for gait planning and control are often overly simplified and handcrafted for particular gait types (e.g. crawl, trot, gallop, e.g.~\cite{bellicoso2018dynamic,fastTraj}). In the worst case, this can limit the versatility of the robot as the models deployed are failing to exploit the full capability of the underlying hardware (e.g.~\cite{bellicoso2018dynamic,ZMP,centroidal_dynamics}). Furthermore, current gait planners are unable to vary key parameters (such as step height) of a gait continuously and online~\cite{towr,melon2021receding}. Instead, contact schedules for discrete gaits are often pre-computed and selected as needed using hierarchical approaches that rely upon a sequence of optimisations, yielding only a narrow operating window which severely limits their response to external disturbances~\cite{towr,melon2021receding}.
The synthesis of gaits which fall outside of this set is typically cumbersome, as approaches that optimise dynamics over gait schedules and footstep lengths or heights are computationally expensive~\cite{towr,melon2021receding}, and often require manual intervention~\cite{bellicoso2018dynamic}. In contrast, the ability to control key gait parameters -- such as cadence, swing height, and full-support duration -- \emph{on-the-fly} would enable a smooth interpolation between dynamic manoeuvres allowing for swift reaction to external stimuli.

Inspired by recent work on a quadruped that achieves a crawl gait via the traversal of a learned latent space~\cite{first-steps}, we approach the challenge of continuous contact-schedule variation from the perspective of learning and traversing a structured latent-space. This is enabled by learning a \emph{generative model} of locomotion data which, in addition to capturing relevant structure in the space, enables the detection and mitigation of disturbances to provide a versatile and robust planning framework. In particular, we train a variational auto-encoder (VAE)~\cite{vae,vae_1} on short sequences of state-space trajectories taken from a single gait type (trot), and predict a set of future states. We show that the resulting latent space has an interpretable structure, lending itself to the generation of a variety of trot styles, depending on how the latent space is traversed. In fact, examining trot trajectories in latent space reveals an oscillatory drive-signal, and we subsequently find that by overwriting this trajectory with a synthetic drive signal, we can \emph{continuously} control the robot's gait properties whilst the robot is executing the motion. Parameters of this drive signal can be mapped explicitly to gait parameters such as cadence, footstep height, and full-stance support duration. 

We illustrate the efficacy of our approach by generating a range of continuously blended trajectories on a real ANYmal quadruped robot -- a medium-sized platform (\SI{35}{\kilo \gram}) standing \SI{0.5}{\meter} tall. While the latent space is learnt using examples only from a specific gait style, our approach is able to synthesise behaviours significantly beyond this training distribution. 

In addition, we leverage our generative approach to both characterise and react to external perturbations. 
A large impulse applied to the robot's base triggers a spike in the Evidence Lower Bound (ELBO) which identifies the disturbance as out of the observed training distribution.
Inspired by \cite{Moyer2006GaitPA}, which states that an increase in cadence is a response to slippages in humans, our planner automatically increases the robot's cadence to aid in counteracting the disturbance and demonstrate a marked improvement in robustness.

To the best of our knowledge, our method is the first which allows continuous variation of the robot's gait characteristics whilst the robot is walking. It provides a versatile and data-driven approach to quadruped locomotion which additionally allows for disturbance detection and recovery. Beyond these contributions, our results are tantalising as the notion of the drive signal is reminiscent of central pattern generators (CPGs) located in the spinal cord and brain stem of vertebrates \cite{steuer2019}. Both CPGs and our synthetic drive signal generate cyclic patterns that effect holistic motor activity and stereotyped rhythmic behaviours such as locomotion.

%% file: 2_related_work.tex
\section{Related Work}

Planning and control for quadruped locomotion has advanced in leaps and bounds in recent years. A seminal work in these areas is \emph{Dynamic Gaits} (DG)~\cite{bellicoso2018dynamic}. DG enables a quadrupedal robot (ANYmal) to execute a wide variety of dynamic gaits (e.g.~trot, pace, lateral walk, jump) with real-time motion planning and control. However, to achieve this impressive range of behaviours, DG provides each gait type with its own contact schedule and utilises an environment-specific footstep planner, ultimately limiting its capability. %

Latent space approaches for planning and control learn useful and typically low-dimensional representations that can be used to control complex dynamics, without relying on known system models. Classic examples include \emph{Deep Variational Bayes Filters} (DVBF) \cite{DVBF} and \emph{Embed to Control} (E2C) \cite{embed2control}. 
\emph{Conditional Neural Movement Primitives} (CNMP) \cite{CNMP} is a more recent latent space approach for robotic arms that generalises between a variety of tasks, such as pick-and-place and obstacle avoidance. 
Other recent works like UPN and PlaNet \cite{UPN,PlaNet} show impressive capabilities in simulation but are yet to be applied to real-world systems, including floating-base robots.

In the quadruped domain, \emph{First Steps}~\cite{first-steps} learns a structured latent-space based on feasible robot \emph{configurations} and defines a set of performance predictors that are used in an optimisation framework to control the robot. In practice, these performance predictors can be viewed as symbolic inputs (e.g.~`left front leg up') but drive the robot in continuous space. Since \emph{First Steps} is trained on static snapshots of robot configurations, it does not learn from observable dynamics and may thus require more explicit structuring of the system than is necessary. Our work addresses this shortcoming and significantly extends this framework to an effective and robust closed-loop planning and control.

The \emph{Motion VAE} (MVAE) \cite{motionVAE} learns to represent dynamic trajectories in a structured latent-space for the locomotion of animated humanoids. This is similar to our emphasis of learning representations for dynamic trajectories in the context of locomotion. However, moving from simulated to real physical systems, as is required for robotic applications, necessitates tackling additional complexities like latency, hard real-time requirements, and actuator dynamics.
In this work, we tackle these challenges and demonstrate that a single gait style contains sufficient richness to learn a structured latent space that can be exploited to manipulate gait characteristics.
Unlike MVAE, our approach does not train on multiple gait styles, despite succeeding in producing them.

%% file: 3_methods_section.tex
\section{Method} \label{section:method}

Our aim is to use unsupervised learning to infer a structured latent-space which facilitates real-time and smooth variation of key gait parameters. 
We conjecture that structure %
can emerge from the exposure of a suitable generative model to a gait with predetermined and constant characteristics such as cadence, swing height, and full-support duration.
We discover a disentangled latent-space where gait parameters are axis-aligned within the structured latent-space.
Periodic trajectories in this space can be decoded leading to smooth robot locomotion as depicted in Fig.~\ref{fig:first_figure}.

\textbf{VAE Architecture:} We train a VAE~\cite{vae,vae_1} to create a structured latent-space using observed dynamic data.
The input $\rmX_k$ at time step $k$ consists of $N$ robot states sampled from simulated trot gaits with constant parameters (e.g. cadence and foot-step height).
These state-space quantities are values we wish either to control or help to infer the gait phase.
These are joint angles; end-effector positions in the base frame; joint torques; contact forces; the base velocity; and the base pose evolution relative to a control frame, which is updated periodically.
These quantities are denoted as ${\rvx_k = [\rvq_k, \mathbf{ee}_k, \mathbf{\tau}_k, \mathbf{\lambda}_k, \mathbf{\dot{c}}_k, \Delta \rvc_k]}$.
Note that velocities and accelerations do not form part of the VAE's input, but are inferred from the input history.
This prevents sensitivity to fast-changing quantities such as recorded joint accelerations during inference.

During deployment on the robot, we encode $\rmX_k$ at the control frequency $f_c$.
However, we construct this input using states $\rvx_k$ sampled at a frequency $f_{enc}$ lower than $f_c$:
\begin{equation}
    \rmX_k = [\rvx^\T_{k-r(N-1)}, \hdots, \rvx^\T_{k-r}, \rvx^\T_k], 
\end{equation}
where $r=f_{c}/f_{\enc}$ is the ratio between the control and encoder frequencies. 
Next, the VAE's decoder output $\hat{\rmX}^{+}_k$ predicts the current robot state $\rvx_k$ as well as $M$ future ones sampled at a frequency of $f_{\dec}=f_c$:
\begin{equation}
    \hat{\rmX}^{+}_k = [\hat{\rvx}^\T_k, \hat{\rvx}^\T_{k+1}, \hdots, \hat{\rvx}^\T_{k+M}] 
\end{equation}

As the \emph{desired-feet-in-contact} $\rvs_k$ is an input to the tracking controller, we also want to predict this at the current time step, and $J$ steps in the future.
Inspired by \emph{First Steps}~\cite{first-steps}, we therefore utilise a feet-in-contact performance predictor $g_{pp}(\rvz_k)$. This is attached to the latent space, and estimates the probability of each foot being in contact:
\begin{align}
    \hat{\rmS}_k &= [\hat{\rvs}^\T_k, ..., \hat{\rvs}^\T_{k+J-1}]^\T 
\end{align}

To command the base twist of the robot, a high-level action $\rva_k$ is utilised.
This represents longitudinal ($x$), lateral ($y$), and yaw ($\theta$) twist in the robot's base frame.
The latent state $\rvz_k$ and the action $\rva_k$ form the input to the decoder.

\textbf{Training the VAE:} 
We train the VAE and performance predictor together. The VAE's training loss is the standard ELBO formulation consisting of a reconstruction loss (mean-squared error) plus the Kullback–Leibler (KL) divergence $\KL$ between the inferred posterior $q(\rvz|\rmX_k)$ and the prior $p(\rvz)$, multiplied by a hyper-parameter $\beta$:
\begin{align}\label{eq:loss_vae}
    \mathcal{L}_{\ELBO} = \MSE(\rmX^{+}_k, \hat{\rmX}^{+}_k) + \beta \KL[q(\rvz|\rmX_k) || p(\rvz)].
\end{align}

These ELBO terms are then summed with the scaled binary cross-entropy loss between the predicted feet in contact and the recorded ones, such that the overall loss is computed as %
\begin{align}\label{eq:loss_full}
    \mathcal{L} = \mathcal{L}_{\ELBO} + \gamma \BCE(\rmS_k, \hat{\rmS}_k)
\end{align}\label{eq:losses_1}
The VAE training loss (Eq.~\ref{eq:loss_vae}), as seen in prior work \cite{betaVAE}, is responsible for any subsequent disentanglement found in the latent space.
The reconstruction error is weighed against the decomposition of the latent space using the hyper-parameter $\beta$.
This constraint encourages an efficient latent representation, containing only the required information for reconstruction, whilst also reducing the channel capacity.
As shown in \cite{betaVAE}, the $\KL$ term used with an isotropic unit Gaussian ($p(\rvz)=\mathcal{N}(\mathbf{0},\rmI)$) encourages conditional independence within $\rvz$.

In our approach, as well as that of \emph{First Steps}~\cite{first-steps}, a structured latent space is encouraged by backpropagating gradients from the performance predictor's loss through to the encoder input.
We hypothesise that useful structuring of this space is inferred from the continuous trajectories used for training, and, in contrast to \emph{First Steps}, no explicit labelling for each stance is required.

\subsection{Investigating the Latent Space}\label{section:methods_structure_latent_space}

Before solving for locomotion trajectories by planning in latent space, we examine the space to see what structure exists within.
Disentangled features are discovered inside.

\begin{figure}[t]
    \centering
    \includegraphics[width=0.85\linewidth]{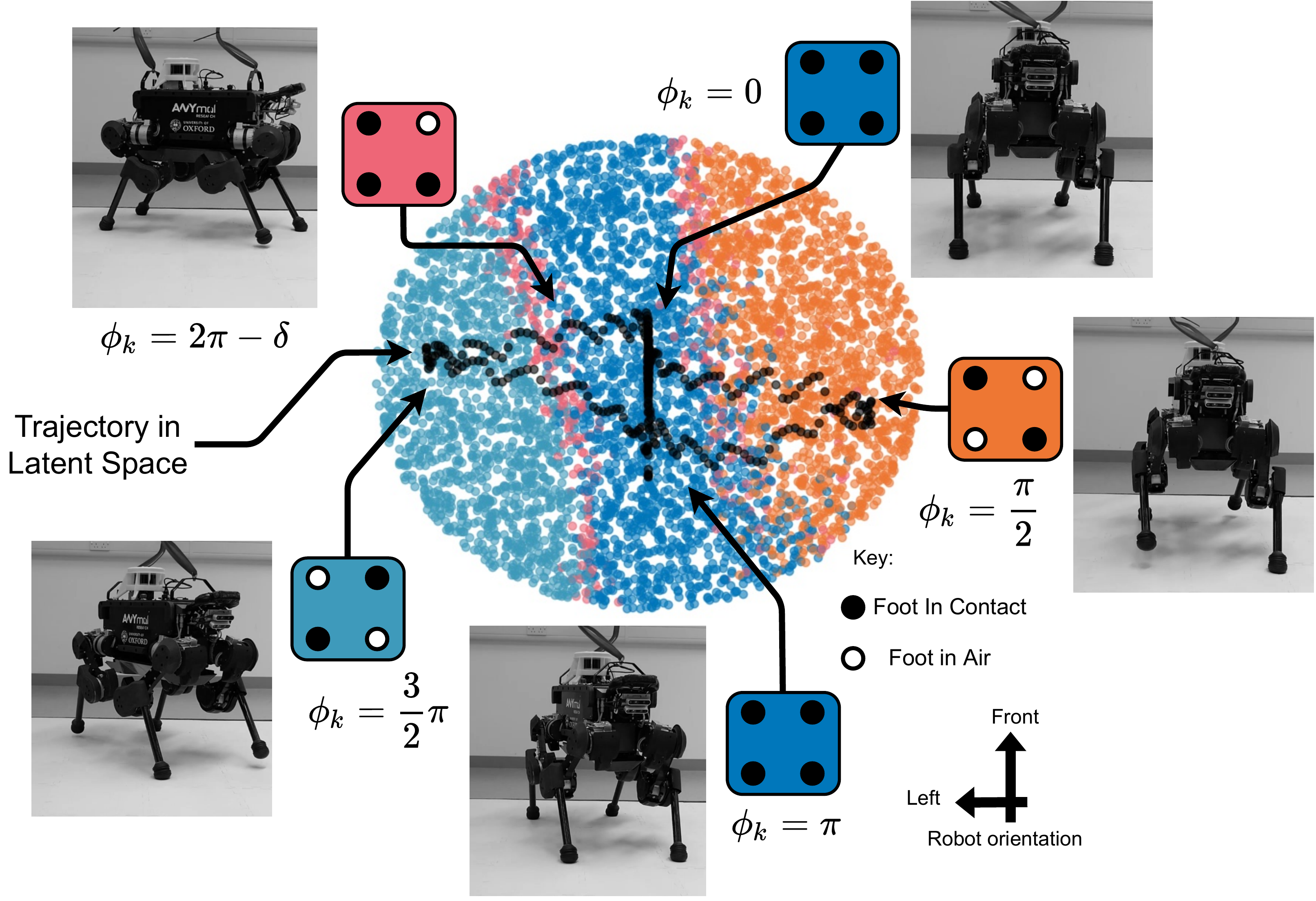}
    \caption{
    A slice through the structured latent space, colour coded to illustrate the ordering and clustering of trot gait stances.
    The horizontal displacement of the latent-space trajectory (black) contributes to the robot's footstep height, whilst the vertical movement relates to the footstep length.
    Snapshots of the robot controlled using the VAE-planner illustrate the inter-play between these two latent dimensions.
    }
    \label{fig:latent-space}
    \vspace{-0.5cm}
\end{figure}

\textbf{Latent Space Structure:}
Fig.~\ref{fig:latent-space} shows samples from the latent space colour-coded by their predicted stance.
For trot there are four stances: the full-support phase, left front and right hind in contact, another support phase and finally right front plus left hind in contact. 
Fig.~\ref{fig:latent-space} reveals that the latent space has emerged clustered by stance and that, due to the ordering of these stances, a periodic trajectory decodes to a trot gait.
This favourable structure is inferred from the continuous trot trajectory input during training.

\textbf{The Effect of Disentanglement:} 
By examining the latent variables, we discover that oscillations injected into just two dimensions in the latent space decode to continuously varying trot trajectories. This result stems from a latent space where variation in footstep length and cadence is aligned along one dimension, while variation in footstep distance lies along another.
The consequence of this disentanglement is that two oscillations (one being $\pi/2$ out of phase with the other) injected into the latent-space is sufficient to produce a broad range of continuously variable trot-trajectories.
Given that other work has tried to explicitly build this structure into locomotion systems~\cite{yang2021fast}, it is important to emphasise that this disentanglement \emph{emerges} in our study as a result of the training paradigm and data.

\subsection{Control Over the Gait Parameters}\label{section:drive-signal}

The disentangled latent-space is exploited such that the cadence, footstep height, and full-support duration can be controlled online.
A specific \emph{drive signal} overwrites a single latent variable, and in this case, we utilise a modified $\sin^3$ oscillation 
with commandable amplitude $A_k$ and phase $\phi_k$:
\begin{align}\label{eq:drive_signal}
  \rvz_{k,d_z} =  A_k \sin^3(\phi_k).
\end{align}
The amplitude $A_k$ controls the foot swing height, and the phase $\phi_k$ governs cadence and support duration. 
To control the robot's swing and stance duration separately, we set the drive signal's time-period $T_k$ and we employ a stance counter $\epsilon_k$.
The time-period $T_k$ is equal to the swing duration, whilst the time that the drive signal is equal to zero is extended by $\epsilon_k$ time steps to introduce a full-support duration of \SI[parse-numbers = false, number-math-rm = \ensuremath]{(\epsilon_k / f_c)}{\second}.
Hence, once both $T_k$ and $\epsilon_k$ are used together, the phase dynamics are:
\begin{align}\label{eq:phase_dynamics}
    \phi_{k+1} = 
    \begin{cases}
      \phi_k & \text{if $\phi_k \bmod \pi = 0$ and $k_{\epsilon} < \epsilon_k$} \\
      \phi_k + {2\pi}/{T_k} & \text{otherwise}
    \end{cases}
\end{align}
and, in tandem, the counter $\epsilon_k$ is updated as:
\begin{align}\label{eq:stance_dynamics}
    k_{\epsilon} \gets 
    \begin{cases}
      k_{\epsilon} + 1 & \text{if $\phi_k \bmod \pi = 0$ and $k_{\epsilon} < \epsilon_k$} \\
      0 & \text{otherwise}
    \end{cases}
\end{align}

\subsection{Planning for Closed-Loop Control}\label{section:closed-loop}

The VAE is fast enough to act as a planner in a closed-loop controller.
Thus, our approach can react to external disturbances and mitigate against real-world effects such as unmodelled dynamics and hardware latency.
For closed-loop control, we begin by encoding a history of robot states from the raw sensor measurements to infer the current gait phase.
We store a buffer of past robot states and sample from this at $f_{enc}$ to create the encoder's input.

With an estimate of the current latent variable, we overwrite latent dimension $d_z$ with the drive signal (see Sec.~\ref{section:drive-signal}).
Next, we employ a second-order Butterworth filter to smooth the latent trajectory and further smooth the locomotion plan.
In essence, the drive signal encourages the decoder to output the next open-loop prediction while the other latent variables infer the gait phase from the raw sensor input.
This process is repeated at the control frequency (\SI{400}{\hertz}).

The latent variable $\rvz_k$ and a desired base twist $\rva_k$ are decoded to produce ${\hat{\rmX}^+_k=g_{\dec}(\rvz_k, \rva_k)}$.
From this, the joint-space trajectory $\hat{\rmQ}_k$, and local base velocity $\hat{\dot{\rmC}}_k$ are extracted and derived or integrated to produce the base and joint positions, velocities and accelerations.
These quantities and the predicted contact schedule are sent to the Whole-Body Controller (WBC)~\cite{WBC}.
The WBC solves a hierarchical optimisation problem to calculate the joint torques which are sent to the actuators.
The series of constraints enforced by the WBC are: contact creation, friction constraints and torque limits.
Next, the WBC applies forward kinematics to the VAE's trajectory to track it in task-space.
Note that the WBC does not compensate for infeasible plans.

\textbf{Disturbance Detection and Response:}
Our approach is able to both detect and react to disturbances. The VAE is trained using canonical feasible trajectories. Therefore, any disturbances are characterised as out of distribution with respect to the training set.
Given the generative nature of our approach, this discrepancy is quantified during operation by the trained model via the Evidence Lower-Bound (ELBO, Eq.~\ref{eq:loss_vae} where $\beta$ is set to one). We will show in the evaluation (Sec.~\ref{section:disturbance_rejetion}) that even a rudimentary response strategy serves to increase the range of disturbance the system can reject.

\begin{figure*}[t]
    \centering
    \includegraphics[width=1.0\textwidth]{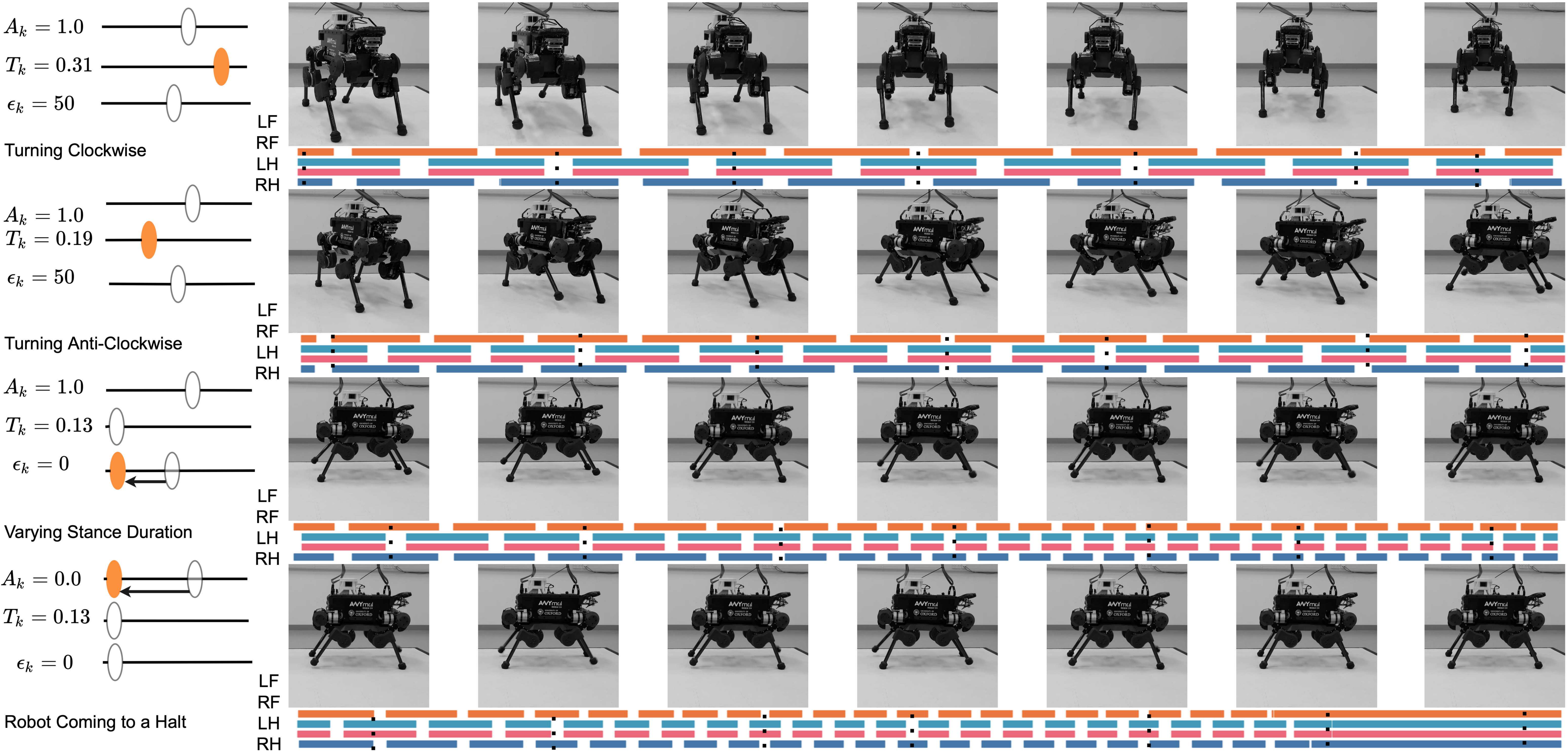}
    \caption{Closed-loop control of the real ANYmal quadruped
    showcasing user-controlled variation of gait parameters on the fly. %
    Coloured rectangles represent the full-stance phase, whilst white space denotes the swing duration.
    The top row shows a trot gait with an introduced quadrupedal stance phase (gait cycle of \SI{0.75}{\second} -- swing \SI{312.5}{\milli\second}, stance \SI{62.5}{\milli\second}; a gait cycle consists of a swing phase for each of the leg pairs).
    Next, the swing duration $T_k$ is reduced using the time-period slider (gait cycle of \SI{0.5}{\second} -- swing \SI{188}{\milli\second}, stance \SI{62.5}{\milli\second}).
    The third row illustrates the effect of reducing the stance duration counter $\epsilon_k$ to produce trot without full-stance phases (gait cycle of \SI{250}{\milli\second} -- swing \SI{125}{\milli\second}, stance \SI{0.0}{\second}).
    Finally, transition into standing occurs when the drive signal amplitude $A_k$ is reduced to zero.
    }
    \label{fig:cadence-comparison}
    \vspace{-0.5cm}
\end{figure*}

%% file: 4_experiments.tex
\section{Experimental Results} \label{section:evaluation}
We demonstrate the capabilities of the VAE as a flexible and robust locomotion-planner deployed for closed-loop control on a real robot.
In doing so, we investigate (i) the structure induced in the latent space (Sec.~\ref{section:embodiment}), (ii) the sensitivity of our approach to variations in key hyper parameters (Sec.~\ref{section:ablation_study}), (iii) to what extent the locomotion parameters can be varied online (Sec.~\ref{section:sliders}), and finally, (iv) the degree to which disturbance detection, coupled with a rudimentary recovery strategy, further increases the robustness of our approach (Sec.~\ref{section:disturbance_rejetion}). 
Please see the following video for an extended set of experiments along with a brief description of our approach (\url{https://youtu.be/2XMyAAe7yDs}).

\subsection{Details on Training and Deployment}\label{section:architecture_details}

Before incorporating our VAE-planner into a closed-loop controller and evaluating the results as deployed on the real ANYmal robot, we discuss the following: firstly, how the dataset is constructed; secondly, how the VAE is trained; and lastly, implementation details for transfer to the real robot.

\textbf{Dataset Generation:}
To train the VAE, we require a set of continuous trot trajectories, and use \emph{Dynamic Gaits} (DG)~\cite{bellicoso2018dynamic} for this.
DG is a hierarchical planning and control framework using a \emph{fixed} contact schedule and predefined footstep heights.
Specifically, swing, full-stance durations and footstep height are set to \SI{0.5}{\second}, \SI{75}{\milli\second} and \SI{0.10}{\meter}.
The components of DG are a footstep planner, a base motion planner and a WBC.
The footstep planner computes the next four steps over the gait period using an inverted-pendulum model.
Using the footstep positions and schedule, the base motion planner computes the base trajectory over a gait period using a centroidal dynamics model~\cite{centroidal_dynamics} constrained with a Zero-Moment Point (ZMP)~\cite{ZMP} criterion.
The WBC is described in Sec.~\ref{section:closed-loop} and converts the task space trajectories to joint torques, positions and velocities, which are sent to the actuators.
The dataset is generated by uniformly sampling desired base twist and executing DG in the \emph{RaiSim} physics simulator~\cite{raisim}.
To improve the fidelity of the simulation, the dynamics of the Series-Elastic Actuators (SEA)~\cite{SEA} are modelled using an \emph{actuator network}~\cite{gangapurwala2020rloc} and accounts for the commanded positions, velocities, feed-forward torques, and low-level PD gains.

\textbf{Architecture Details:}
In all our experiments, the VAE's encoder, decoder and stance performance predictor have two hidden layers and widths of 256 units, using ELU non-linearities~\cite{elu}. 
The encoder input is created using $N=80$ robot states sampled at \SI{200}{\hertz} -- representing a history of \SI{0.4}{\second} -- from the encoder input, which is of size $5120$ units.
The input is compressed via a latent space of $125$ units which is concatenated with an action of $3$ units.
Next, the decoder outputs the current state and the next $M=19$ robot states at the control frequency of \SI{400}{\hertz} (preview horizon \SI{47.5}{\milli\second}, output size: $1216$ units). 
The performance predictor predicts the current feet in contact and two future states.
Finally, hyper-parameters used for training are $\beta=1.0$, $\gamma=0.5$, with a learning rate of \num{1e-3} using the Adam optimiser.
Training is terminated after \num{1e+6} gradient steps.

\begin{figure*}[t]
    \centering
    \begin{subfigure}[t]{0.7\textwidth}
    \includegraphics[width=1.0\textwidth]{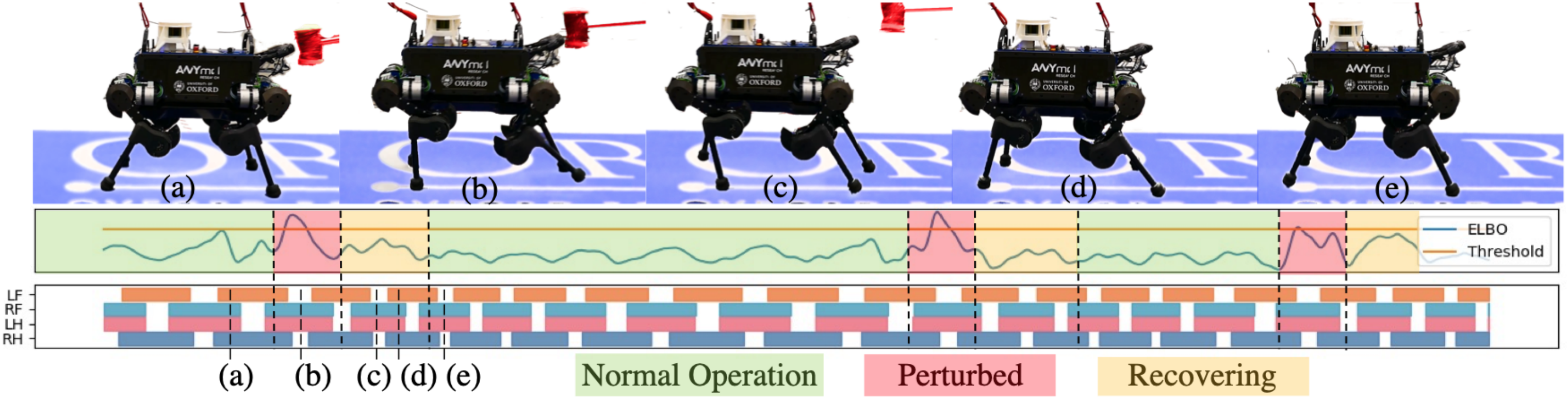}
    \end{subfigure}~~~~~~~~~~~~ %
    \begin{subfigure}[t]{0.29\textwidth}
    \includegraphics[width=0.6\textwidth]{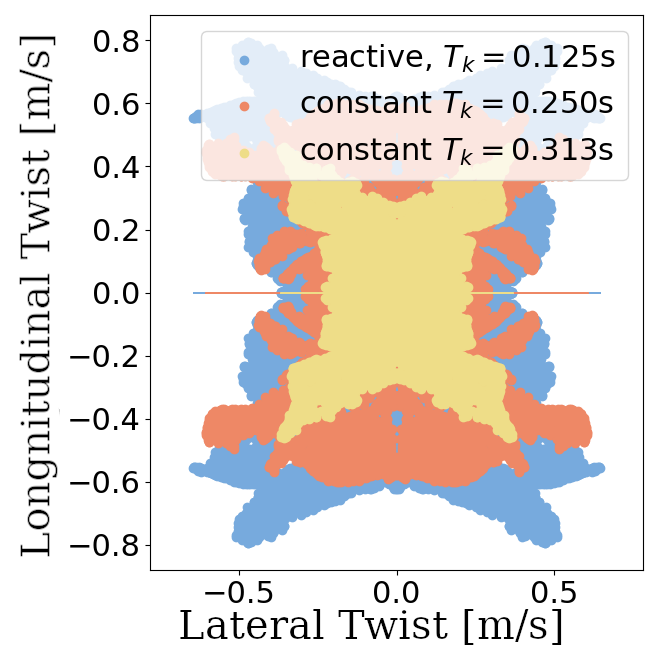}
    \end{subfigure}
    \caption{Push recovery following a shove to the base. A disturbance to the robot causes the ELBO of the form in Eq.~\ref{eq:loss_vae} to rise above a predetermined threshold (red areas). This identifies a disturbance and triggers an increase in cadence from \SI{250}{\milli\second} to \SI{125}{\milli\second} as a rudimentary response. This reduction in the swing duration is visible in the contact schedule.
    The right sub-plot shows how the range of rejectable base-velocity disturbances increases when automatically modulating the cadence.
    To appreciate the results fully, please see the following video \url{https://youtu.be/2XMyAAe7yDs}.}
    \vspace{-0.5cm}
    \label{fig:push_recovery}
\end{figure*}

\textbf{Domain Transfer:} 
To achieve domain transfer and deployment on the real robot, some modifications are required.
The contact forces are artificially set to zero when the robot measures no contact.
As mentioned in Sec.~\ref{section:closed-loop}, a Butterworth filter with a cutoff frequency of \SI{10}{\hertz} smooths the latent trajectory.
Due to the strict computation budget for real-time control, inference times for the VAE are restricted to at most \SI{1}{\milli\second}, imposing constraints on the capacity of the model (see Sec.~\ref{section:ablation_study}).
Our largest model takes approximately \SI{1}{\milli\second} for the VAE computation, which is roughly equal to the computation time of the WBC.

\subsection{Structure Induced In The Latent Space}\label{section:embodiment}

The latent space is inspected to discover what structure exists and if any locomotion properties are disentangled within.
Fig.~\ref{fig:latent-space} depicts the latent space and is plotted such that the drive signal dimension is along the horizontal, and trot signal dimension is aligned with the vertical.
In addition, the stances (coloured areas) are distinctly clustered in sequence of the trot gait, whilst the user-controlled latent-space trajectory (black) moves through the space in a figure of eight visiting each stance cluster sequentially.
See Sec.~\ref{section:methods_structure_latent_space} for detailed discussions.

\subsection{Sensitivity To Hyper Parameters}\label{section:ablation_study}
The size of the latent dimension is reduced from $125$ to a minimum of $6$ in strides of $32$ units.
The latent dimension of $6$ units deployed in simulation and the real robot produces trajectories which are stable and well tracked by the WBC.
Next, the input history is reduced from \SI{0.4}{\second} to \SI{0.3}{\second} sampled at \SI{200}{\hertz}, resulting in poor open-loop performance. 
Since the gait's swing duration in training lasts \SI{0.45}{\second}, an input history long enough to capture this is necessary to infer the gait phase.
Finally, the VAE's width is reduced in 32 unit increments to reduce the channel capacity until the VAE's trajectories cause the robot to become unstable in simulation.
This limit is reached at a width of $128$ units, corresponding to a reduction in channel capacity by \SI{52.8}{\percent}. %
We verified these models on the robot showcasing that a minimal model requires a latent space of $6$ units, an input history of \SI{0.4}{\second} sampled at \SI{200}{\hertz}, and a hidden layer size of $128$ neurons.

\subsection{Varying Locomotion Parameters Online} \label{section:sliders}

We leverage the disentangled latent-space to smoothly transition between gait parameters whilst the robot is walking.
Crucially, cadence, stance duration, and footstep height can be varied during any phase of the gait by modulating the drive signal's parameters (see Sec.~\ref{section:drive-signal}).
This results in operating modes which vary from those seen during training.
Examples of walking motion using the VAE-planner on the real robot are shown in Fig.~\ref{fig:cadence-comparison}.

\textbf{Swing Duration:} 
The swing duration is varied over a large operating window on the ANYmal robot.
This begins with a swing time-period starting at \SI{312.5}{\milli\second}, and is smoothly varied until the swing duration reaches \SI{125}{\milli\second}, effecting a faster step rate.
In parallel, we alter the robot's heading, demonstrating the independence of the action and the latent-space dynamics.
Specifically, the top row of Fig.~\ref{fig:cadence-comparison} shows the nominal swing duration of \SI{312.5}{\milli\second} as the robot turns clockwise, tracking a constant angular velocity command.
Next, we demand a slightly faster swing \SI{188}{\milli\second} and a constant angular velocity anti-clockwise, before transitioning to the fastest swing (\SI{125}{\milli\second}) in the third row.
Here, the coloured contact schedule captures the changes in swing duration as it occurs in real-time.

\textbf{Stance Duration:} Following a successful reduction in cadence, the stance duration is reduced and the robot transitions into a trot with negligible full-stance phase: $\epsilon_k=0$.
Trot gaits with little to no full-support phase are particularly challenging manoeuvres for the system in general, as there is reduced control authority to correct for accrued base pose error.
During the swing phase of this gait style, only feet across the diagonal are in contact resulting in a line contact limiting the robot's ability to steady its base.
The transition to a negligible full-stance phase is captured in the third row, where the coloured stance duration reduces in length.

\textbf{Footstep Height:} We vary the amplitude of the drive signal smoothly to zero as is seen in the bottom row (Fig.~\ref{fig:cadence-comparison}): %
The footstep height reduces to zero as the white-space in the contact schedule disappears and the robot remains standing.
Beyond versatility, e.g. to increase swing heights to overcome irregular ground height, this capability further enables a safe, smooth and natural transition into and out of the VAE control mode (i.e. to start and come to a halt).

\vspace{-0.3em}
\subsection{Disturbance Detection and Recovery}\label{section:disturbance_rejetion}

The ELBO is used to detect disturbances as it is a lower bound for the evidence of a sample given a particular distribution.
The distribution in question is that learned over the training data from DG. 
Therefore, any motions which deviate from this due to a perturbation cause a large negative spike in the ELBO value.
For simplicity we use ELBO in the form of Eq.~\ref{eq:loss_vae} meaning that large positive values are a result of a perturbation (see red areas in Fig.~\ref{fig:push_recovery}).

The VAE-planner is able to reject a wide range of impulses applied to the robot’s base. 
This operating window is enlarged by increasing the robot’s cadence once a disturbance is detected, see Fig.~\ref{fig:push_recovery}.
This is a rudimentary response inspired by \cite{Moyer2006GaitPA}, which reports that humans increase their cadence to recover from slippage.
The cadence is only increased if the ELBO value rises above a pre-selected threshold, below which is considered normal conditions.
This value is ${11.0}$ and found by operating the robot for a few minutes and recording the ELBO.
The nominal cadence is set to \SI{250}{\milli\second} and halves to \SI{125}{\milli\second} for \SI{1.50}{\second} when the threshold is surpassed.

Fig.~\ref{fig:push_recovery} depicts the ELBO trace for three push events along with the robot's contact schedule. 
The widths of the white spaces in the contact schedule half as the cadence increases to mitigate the disturbance.
The robot images above this are snapshots taken from the first push and show the robot's recovery.
The robot successfully recovers and this usually requires between three and four steps.

%% file: 5_conclusions.tex
\vspace{-0.1em}
\section{Conclusion}\label{section:conclusion}
We present a robust and flexible approach for locomotion planning via traversal of a structured latent-space.
We utilise a deep generative model to capture features from locomotion data, and enable detection and mitigation of disturbances.
The resulting latent-space is disentangled such that key locomotion features are automatically discovered from a single style of trot gait.
This disentanglement is exploited using an oscillatory drive-signal, where the amplitude and phase directly control the gait parameters, namely the cadence, swing height, and full-support duration.
Once deployed, the ease with which modulation of the drive signal gives rise to seamless interpolation between gait parameters is demonstrated.
Utilising a generative model affords detection of disturbances as out of the distribution seen during training.
Although the VAE-planner is able to reject a broad range of impulses, this window is broadened by increasing the cadence once the disturbance is detected -- a rudimentary response inspired by human locomotion during slippage~\cite{Moyer2006GaitPA}.

%% file: 6_appendix.tex
\vspace{-0.3em}
\section*{Acknowledgements}
The authors acknowledge the use of the SCAN facility, %
and thank Oliver Groth for useful discussions.